\documentclass[conference]{IEEEtran}
\IEEEoverridecommandlockouts
\usepackage{cite}
\usepackage{hyperref}
\usepackage{amsmath,amssymb,amsfonts}
\usepackage{algorithmic}
\usepackage{graphicx}
\usepackage[normalem]{ulem}
\usepackage{textcomp}
\usepackage{multirow}
\usepackage{soul}
\usepackage{xcolor}
\def\BibTeX{{\rm B\kern-.05em{\sc i\kern-.025em b}\kern-.08em
    T\kern-.1667em\lower.7ex\hbox{E}\kern-.125emX}}
\begin{document}

\title{Cumulative Hazard Function Based Efficient Multivariate Temporal Point Process Learning}
\author{\IEEEauthorblockN{Bingqing Liu}
\IEEEauthorblockA{
Academy of Mathematics and Systems Science, CAS\\
University of Chinese Academy of Science, Beijing, China\\
liubingqing20@mails.ucas.ac.cn}
}

\maketitle

\begin{abstract}
Most existing temporal point process models are characterized by conditional intensity function. These models often require numerical approximation methods for likelihood evaluation, which potentially hurts their performance. By directly modelling the integral of the intensity function, i.e., the cumulative hazard function (CHF), the likelihood can be evaluated accurately, making it a promising approach. However, existing CHF-based methods are not well-defined, i.e., the mathematical constraints of CHF are not completely satisfied, leading to untrustworthy results. For multivariate temporal point process, most existing methods model intensity (or density, etc.) functions for each variate, limiting the scalability. In this paper, we explore using neural networks to model a flexible but well-defined CHF and learning the multivariate temporal point process with low parameter complexity. Experimental results on six datasets show that the proposed model achieves the state-of-the-art performance on data fitting and event prediction tasks while having significantly fewer parameters and memory usage than the strong competitors. The source code and data can be obtained from \href{https://github.com/lbq8942/NPP}{https://github.com/lbq8942/NPP}.
\end{abstract}


\section{Introduction}
Event sequences in continuous time space are ubiquitous around us - hospital visits \cite{Mavroudeas2021PredictiveMF,Choi2015ConstructingDN}, tweets \cite{Yao2016TweetTG}, financial transactions \cite{bacry2015hawkes}, earthquakes \cite{van2012earthquake}, human activities \cite{Zhou2013LearningSI,farajtabar2015coevolve,Liu2023LinkAwareLP}, to name a few. Beyond mere timestamps, these sequences usually also carry rich information about the types of events. Inside the event sequence, various types of asynchronous events often exhibit complex hidden dynamics responsible for their generation \cite{Zhang2021LearningNP,Ding2023CNTPPLC,Zhang2021NeuralRI}. Diving into the dynamics, temporal point process (TPP) \cite{daley2003introduction,daley2007introduction} has been a popular and principled tool for event sequence modelling, through the lens of which, these sequences can be interpreted as realizations of an underlying stochastic process. 

Traditional TPPs often use over-simplified historical dependencies \cite{du2016recurrent,waghmare2022modeling,mei2017neural}. For example, Poisson process \cite{kingman1992poisson} assumes the future events are independent of the past events. With the development of deep learning, the so-called
neural temporal point process (NTPP) models \cite{shchur2021neural,Xue2023EasyTPPTO,lin2021empirical,Lin2021ExtensiveDT,yan2019recent} have been intensively devised for more flexible historical dependencies modeling. NTPP is typically trained by maximizing the likelihood \cite{mei2017neural}, which consists of two terms, the intensity and its integral (cumulative hazard function, CHF). Most existing NTPPs are modeled by conditional intensity functions \cite{mei2017neural,soen2021unipoint,zuo2020transformer,zhang2020self}, by which the intensity term in the likelihood can be obtained directly. However, due to the high non-linearity of neural networks, the integral term is usually intractable. As a result, approximation methods are required, e.g., Monte Carlo algorithm, trapezoidal rule, Simpson's rule and etc \cite{mei2017neural,zuo2020transformer}. Not only is the approximation inefficient, but also the inaccurate likelihood evaluation can potentially harm the training and finally the testing performance. FullyNN \cite{omi2019fully} pioneered the CHF-based modelling, by which the integral term is directly in hand, and the intensity can be obtained by taking its derivative. Not like integration, the derivation can be exactly calculated. However, FullyNN fails to define a valid CHF, the unfulfilled mathematical constraints make the achieved performance untrustworthy \cite{shchur2019intensity}. Moreover, FullyNN defines a bounded CHF, limiting its flexibility. 

In the realm of multivariate temporal point process \cite{Cox2018MultivariatePP}, scalability remains a neglected aspect that deserves greater attention. Prior studies usually learn intensity (or density, etc.) functions individually for each variate (event type) \cite{zuo2020transformer,waghmare2022modeling,soen2021unipoint}, which can be prohibitively expensive when dealing with real-world event sequences that encompass numerous event types. In fact, the event observations are so sparse that there's no need seeing the entire view for all event types. 

The contributions of this paper are summarised
as follows:
\begin{itemize}
    \item We present a flexible and well-defined cumulative hazard function network and incorporate it into a parameter-efficient multivariate temporal point process.
    \item We show that the proposed model performs better
in data fitting and predictive tasks for multiple datasets.
    \item We show that our model has lower parameter complexity and memory footprint, with up to 100 $\times$ parameter and memory reduction comparing to state-of-the-art baselines. 
\end{itemize}
 
\section{Preliminaries and Related Work}
Before diving into details, we first introduce some preliminaries to facilitate the presentation of our approach.
\subsection{Temporal Point Process}
The dynamics of a TPP can be represented as a counting process $N(t)$ which counts the number of events that happened until time $t$. A more common definition is by its conditional intensity function:
\begin{equation}
\label{eq:tppdef}
	\begin{aligned}
\lambda(t|\mathcal{H}_{t})&=\lim_{\Delta t\to 0}\frac{\mathbb{E}(N(t+\Delta t)-N(t)|\mathcal{H}_{t})}{\Delta t}
	\end{aligned}
\end{equation}
which we interpret as: for a short time window $[t, t + \Delta t)$, $\lambda(t)$ represents the rate for the occurrence of a new event conditioned on the past events  $\mathcal{H}_{t}=\{t_i|t_i<t\}$. Given the intensity, the density can be derived as:
\begin{equation}
\label{eq:intpdf}
	\begin{aligned}
P_t(t|\mathcal{H}_{t})&=\lambda(t|\mathcal{H}_{t}) \exp(-\int_{t_{max}}^{t} \lambda(s|\mathcal{H}_{s})ds) 
	\end{aligned}
\end{equation}
where $t_{max}$ is the last event time in $\mathcal{H}_{t}$. The conditional intensity function has
played a central role in point processes and many popular
processes differ on how it is parameterized. Classical Poisson process \cite{kingman1992poisson} assumes the future events are independent of the history and defines the conditional intensity function as a constant. Hawkes process (self-exciting process) \cite{hawkes1971spectra} captures the excitation effects from the past events by definition $\lambda(t|\mathcal{H}_{t})=\mu+\sum_{t_i} \alpha \exp(-\beta (t-t_i))$. In self-correcting process \cite{isham1979self}, the past events have inhibition effects on the future events, defined by $\lambda(t|\mathcal{H}_{t})=\exp(\mu t-\sum_{t_i} \alpha)$. Traditional point processes make strong assumptions about the latent dynamics, thus having limited expressive power \cite{du2016recurrent}.
\subsection{Neural Point Process}
An important aspect for temporal point process is to model the historical dependencies \cite{du2016recurrent,dash2022learning,zuo2020transformer}. To be more flexible, neural networks are introduced to encode the events history, such as recurrent neural network (RNN) and Transformer \cite{vaswani2017attention,Gu2021AttentiveNP}. The neural encoder maps a historical event sequence to a low-dimensional vector, based on which conditional intensity  function (or other functions like the density function \cite{shchur2019intensity}) is defined. RMTPP \cite{du2016recurrent} uses RNN to encode the past events as vector $h$ and defines $\lambda(t|\mathcal{H}_{t})=\exp(v^Th+\alpha(t-t_{max})+b)$, where $t_{max}$ is the last event time in $\mathcal{H}_{t}$. Some other works have a similar definition \cite{zuo2020transformer,zhang2020self,upadhyay2018deep}. These models are not flexible because they actually specified a monotonic intensity function.  To enhance the flexibility, NHP \cite{mei2017neural} proposes a novel RNN architecture - Continuous-time LSTM to model a time-varying hidden vector $h(\cdot)$ and defines the intensity function as $\lambda(t|\mathcal{H}_{t})=\sigma(v^Th(t-t_{max}))$, where $\sigma$ is an activation function to ensure the intensity is non-negative. UNIPoint \cite{soen2021unipoint} fits a complex intensity function by adding up a bunch of parametric simple functions, specifically $\lambda(t|\mathcal{H}_{t})=\sigma(\sum_{c=1}^C \alpha_c \exp(\beta_c(t-t_{max})))$, where $\alpha_c,\beta_c$ are parameters learned from history vector $h$. Theoretically elegant, UNIPoint is proven to have the universal approximation property. While these models offer considerable flexibility, they come with a tradeoff for closed-form likelihood evaluation. As a result, approximation methods such as Monte Carlo are required, leading to inaccuracy and inefficiency. To overcome this disadvantage, FullyNN \cite{omi2019fully} proposes to directly model the integral of the intensity function (cumulative hazard function). Technically, FullyNN uses a neural network with positive parameters \cite{Sill1997MonotonicN,Chilinski2018NeuralLV,rindt2022survival} to satisfy the monotonic constraint of cumulative hazard function. In FullyNN, the likelihood can be computed accurately and the flexibility is guaranteed by the neural network. Beyond the monotonicity, there actually exist some other mathematical constraints for cumulative hazard function, which however, are not fulfilled by FullyNN. This makes the performance obtained from FullyNN  implausible. LNM \cite{shchur2019intensity} proposes to  model the probability density function by mixture log-normal distribution, which is able to approximate any complex distribution. The major drawback of LNM is its overfitting problem, which requires variance clipping technique to ensure numerically stable training and inference. 

Multi-type event sequences are ubiquitous across applications, which can be correspondingly characterized by multivariate temporal point process \cite{Cox2018MultivariatePP}. Most existing methods define intensity functions \cite{mei2017neural,soen2021unipoint} $\lambda_k(t|\theta_k,\mathcal{H}_{t})$ (or cumulative hazard functions \cite{omi2019fully}, probability density functions \cite{shchur2019intensity, waghmare2022modeling, liu2024interpretable}, etc.) for each event type $k$, with different types of intensity functions (or other functions) having the same parametric form but different parameters $\theta_k$. By increasing the number of parameters in $\theta_k$, the model can indeed exhibit greater flexibility. However, this tends to be prohibitively expensive in scenarios involving numerous event types. RMTPP \cite{du2016recurrent} scales well but has inferior performance for it only defines a simple intensity function and doesn't consider the inter-dependence between time and event type.
\section{Method}
\subsection{Multivariate Temporal Point Process}
\label{subsec:mtpp}
Given an observed multi-type event sequence $S=\{(m_1,t_1),\cdots,(m_N,t_N)\}$, where $m\in \{1,\cdots, M\},t\in \mathbb{R}^+$ denote the event type and time, event types are categorical and event times are strictly increasing in continuous space, multivariate temporal point process (MTPP) is essentially modelling the conditional joint distribution $P(m,t|\mathcal{H}_t)$, where history $\mathcal{H}_t=\{(m_i,t_i)|t_i<t\}$. Naturally, MTPP model can be learned by maximizing the log-likelihood below:
\begin{equation}
\label{eq:obj1}
	\begin{aligned}
\mathcal{L}_l&=\sum_{i=1}^{N}\log P(m_i,t_i|\mathcal{H}_{t_i})
	\end{aligned}
\end{equation}
Given condition $\mathcal{H}_{t_i}$, we can equivalently model the inter-event time $\tau_i=t_i-t_{i-1}$. For simplicity, we will omit the writing of history condition $\mathcal{H}_{t_i}$ from now on for we are always modelling conditional distributions throughout this paper. A widely adopted joint distribution modelling strategy is to characterize the intensity functions, i.e.,
\begin{equation}
\label{eq:obj2}
	\begin{aligned}
&P(m_i,\tau_i)= P_t(\tau_i)P_m(m_i|\tau_i)\\
&=[\lambda(\tau_i) \exp(-\int_{0}^{\tau_i} \lambda(s)ds) ][\frac{\lambda_{m_i}(\tau_i)}{\lambda(\tau_i)}]\\
&\lambda(\tau_i)=\lambda_{1}(\tau_i)+\cdots+\lambda_{M}(\tau_i)
	\end{aligned}
\end{equation}
where $P_t(\cdot)$ is probability density function defined on $\mathbb{R}^+$, $P_m(\cdot)$ is a $M\mbox{-}$categorical distribution and $\lambda(\cdot)$ is intensity function. In this framework, all intensity functions $\{\lambda_m(\cdot)\}_{m=1}^M$ are learning objectives, which has a high model complexity especially when $M$ is particularly large. Moreover, the log-likelihood contains an integral with respect to the intensity function, which is always intractable and requires approximation methods such as Monte Carlo in practice, leading to inaccurate and inefficient model learning. 

To overcome these shortcomings, we propose to model the joint distribution by only learning the total intensity function and the time-dependent type distribution, i.e.,
\begin{equation}
\label{eq:obj3}
	\begin{aligned}
P(m_i,\tau_i)=\lambda(\tau_i) \exp(-\int_{0}^{\tau_i}\lambda(s)ds)   P_m(m_i|\tau_i)\\
	\end{aligned}
\end{equation}
As a comparison, the aforementioned framework  \eqref{eq:obj2} needs learning $M$ continuous variables $\{\lambda_m(\cdot)\}_{m=1}^M$ while we only have one continuous variable $\lambda(\cdot)$ and one discrete variable $P_m(\cdot|\tau_i)$, which greatly reduces the learning complexity. To avoid the integral computation in \eqref{eq:obj3}, we borrow ideas from FullyNN \cite{omi2019fully}: we directly model the integral function $\int_{0}^{\tau}\lambda(s)ds$, named cumulative hazard function (CHF). Accordingly, the intensity function can be derived through gradient calculation, i.e., $\partial \int_{0}^{\tau}\lambda(s)ds/\partial \tau$, which can be automatically and exactly calculated by existing deep learning library (e.g., Pytorch). As a result, the proposed method can evaluate the likelihood accurately and efficiently.

\subsection{Model Architecture}
To model the joint distribution $P(m_{i+1},\tau_{i+1})$, the first step is to encode the past events $\mathcal{H}_{t_{i+1}}$ into a low-dimensional dense semantic vector $h_{i}$. Commonly used event sequence encoder includes recurrent neural network (RNN \cite{chung2014empirical}) and Transformer \cite{vaswani2017attention}. In this paper, we choose RNN as our encoder for it has competitive performance but fewer parameters. RNN recursively updates the history vector $h$ by digesting new coming events, which can be formulated as follows,
\begin{equation}
\label{eq:rnnup}
	\begin{aligned}
h_i=update(h_{i-1},e_i)\\
	\end{aligned}
\end{equation}
where $e_i \in \mathbb{R}^{d_m+1}$ is the embedding of the $i\mbox{-}th$ event $(m_i,t_i)$, which is specified as the concatenation of the trainable type embedding $\boldsymbol{m_i}\in \mathbb{R}^{d_m}$ and the inter-event time $\tau_i$, telling the encoder which type of event happens and how long has it been since the last event. The operation $update$ in \eqref{eq:rnnup} in vanilla RNN is formulated as follows,
\begin{equation}
\label{eq:vanillarnn}
	\begin{aligned}
h_i=tanh (W_h(h_{i-1}||e_i)+b_h)\\
	\end{aligned}
\end{equation}
where $||$ denotes the concatenation of two vectors, $tanh$ is the activation function for non-linearity, weight matrix $W_h \in \mathbb{R}^{d\times (d+d_m+1)}$ ($d$ is the hidden dimension for RNN) and bias vector $b_h \in \mathbb{R}^{d}$ are trainable parameters.  More complicated $update$ for \eqref{eq:rnnup}, e.g., GRU \cite{chung2014empirical} and LSTM \cite{hochreiter1997long}, can also be used for potentially better performance.

The second step is to model the cumulative hazard function $\int_{0}^{\tau}\lambda(s)ds$ and the time-dependent type distribution  $P_m(m|\tau)$ in \eqref{eq:obj3}. The type distribution is categorical, conditioned on past events $\mathcal{H}_{t_i}$ and the occurrence time. Thus, we model the type distribution as a $M\mbox{-}$dimensional vector that has history- and time-varying elements. Specifically, we use a two-layer neural network, i.e.,
\begin{equation}
\label{eq:typepred1}
	\begin{aligned}
P_m(m|\tau)=softmax(W_m^2y_m^2+b_m^2)\\
y_m^2=relu (W_m^1(h_{i}||\tau)+b_m^1)\\
	\end{aligned}
\end{equation}
where weight matrices and biases $W_m^1\in \mathbb{R}^{d\times(d+1)},W_m^2\in \mathbb{R}^{M\times d},b_m^1\in \mathbb{R}^{d},b_m^2\in \mathbb{R}^{M}$ are all trainable parameters, $relu$ is the rectified linear unit, defined by $max(0,\cdot)$, $softmax$ is a normalizing function that makes an arbitrary $M\mbox{-}$dimensional vector a categorical probability, with each element's exponent divided by the exponential sum.

The cumulative hazard function $\phi(\tau)=\int_{0}^{\tau}\lambda(s)ds$ is monotonic  with respect to the inter-event time $\tau$ since the intensity function is non-negative. FullyNN \cite{omi2019fully} simulates the CHF  by a neural network where the neural weights are all positive and the activation functions are all monotonic. However, neural network defined that way doesn't meet the requisite $\phi(0)=0$, making it a flawed design. To repair it, we try not posing any extra restrictions or doing any modifications on the network architecture so as to not harm the expressiveness. For this sake, we reproduce a well-defined CHF network by dynamically subtracting the initial bias $\phi(0)$. We say the bias is dynamic because it's history-dependent rather than a fixed something (e.g., network architecture and its parameters). The whole cumulative hazard function network (three-layer for example) is then defined as, 
\begin{equation}
\label{eq:chf1}
	\begin{aligned}
\phi(\tau)&=f(\tau)-f(0)\\
f(\tau)&=v_t^Tz_t^3+b_t^3\\
z_t^3&=\sigma_t(W_t^2z_t^2+b_t^2)\\
z_t^2&=\sigma_t(W_t^1(h_i||\tau)+b_t^1)\\
	\end{aligned}
\end{equation}
where neural weights $W_t^2\in \mathbb{R}^{d\times d},v_t\in \mathbb{R}^{d}$ in the second and third layer are constrained to be positive. For the weight matrix $W_t^1\in \mathbb{R}^{d\times (d+1)}$ in the first layer, the last column vector that acts on $t$ is also required to be positive. For the remaining parameters in $W_t^1$ and biases $b_t^1\in \mathbb{R}^{d},b_t^2\in \mathbb{R}^{d},b_t^3$, we don't pose the positive constraint. In addition to the constraints on parameters, activation function $\sigma_t$ is supposed to be monotonic to ensure the monotonicity of the whole network. We use $tanh$ function for this purpose. We have also
tried some other commonly used monotonic activation functions such as $relu$ and $softplus$ but empirically see inferior performance, which we will show in the experiment. So far, we have designed a well-defined CHF network by \eqref{eq:chf1}.
\begin{figure*}
\centering 
\includegraphics[width=0.85\textwidth]{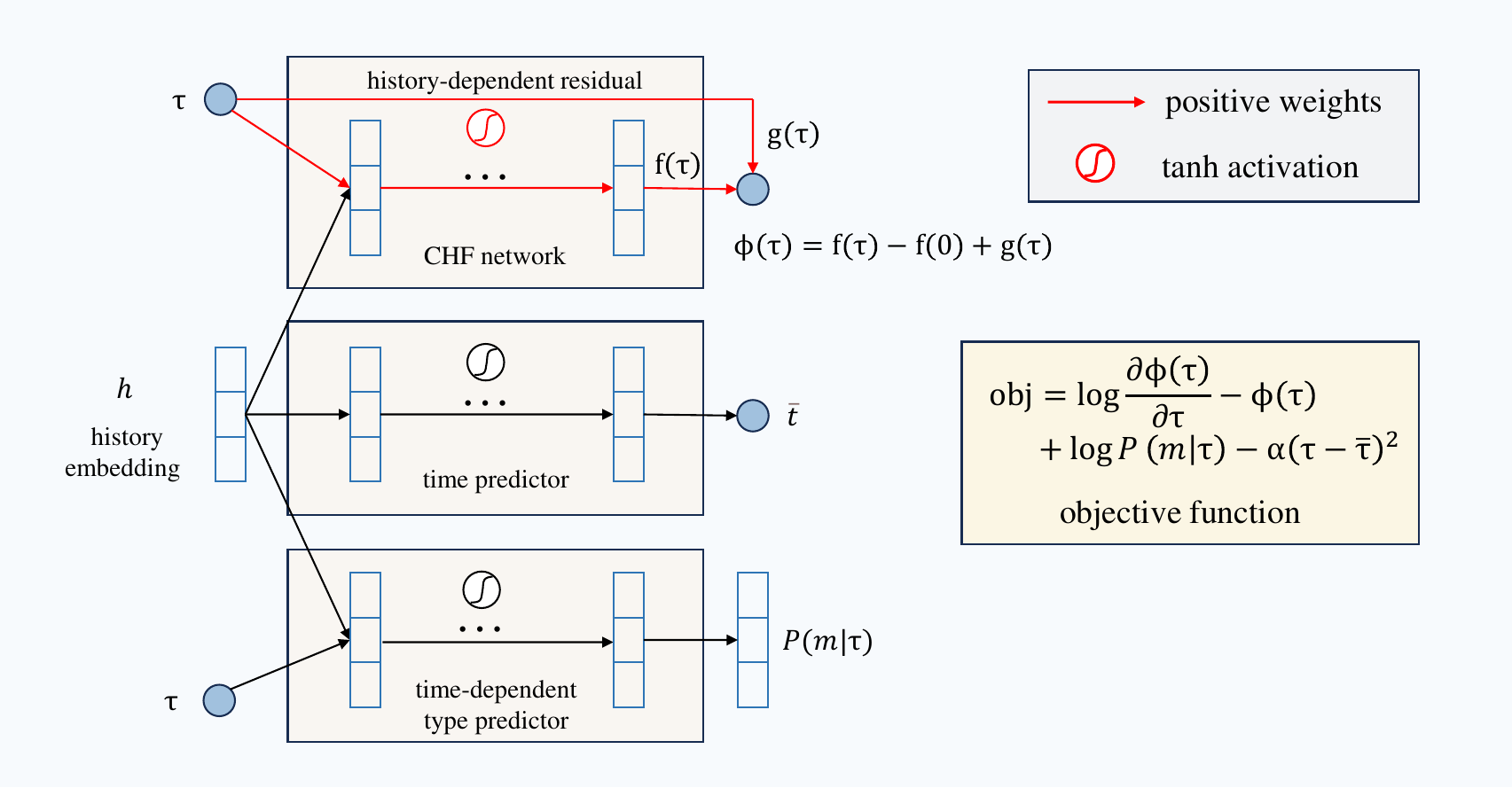}
\caption{The proposed cumulative hazard function (CHF) based multivariate temporal point process model, including three modules: the CHF network, the time predictor and the time-dependent type predictor. The CHF network is to reproduce a well-defined CHF so as to capture the temporal dynamics, the time predictor is for time prediction, and the time-dependent type predictor is to capture the dynamics of event types and also used for type prediction.}
\label{fig:model}
\end{figure*}

The activation function $tanh$ is highly non-linear and can activate both positive and negative numbers. However, $tanh$ is a bounded function, with the gradient vanished when saturated.  The gradient vanishing problem makes the intensity function converges to zero after a period of time, which may conflict with reality. In fact, many events happen not only because of the exogenous factors (excitation or inhibition from other events) but also by endogenous ones (occur spontaneously without external intervention), which is called base intensity and usually modelled as a constant in literature \cite{hawkes1971spectra, du2016recurrent, zuo2020transformer}. To capture this pattern, we enhance the flexibility of our CHF network by adding a history-dependent linear residual. Specifically, we modify $\phi(\tau)$ in \eqref{eq:chf1} as follows,
\begin{equation}
\label{eq:chf2}
	\begin{aligned}
\phi(\tau)&=f(\tau)-f(0)+g(\tau)\\
g(\tau)&=\gamma \tau, \gamma=\sigma_r(v_{\gamma}^Th_i+b_{\gamma})\\
	\end{aligned}
\end{equation}
where $v_{\gamma} \in \mathbb{R}^d, b_{\gamma}$ are parameters to be trained and $\sigma_r$ is the $relu$ activation function, which is non-negative and can ensure the monotonicity of the CHF network $\phi(\tau)$. With the added residual, the intensity function $\partial \phi(\tau)/\partial \tau$ now converges to the base intensity $\gamma$. Notably, the residual can adapt itself to the history, gaining more flexibility. If $\gamma$ is learned to be 0, then the CHF network defined in \eqref{eq:chf2} degrades  to that in \eqref{eq:chf1}.

When doing prediction on the $(i+1)\mbox{-}th$ event, the proposed model should first predict the event time and then predict the event type according to \eqref{eq:obj3}. The event time is predicted via the CHF network $\phi(\tau)$ and the event type is predicted through the time-dependent type predictor $P_m(m|\tau)$. Specifically, if we use the expectation $\int_{0}^{\infty}sP_t(s)ds$ as the predicted time $\hat{\tau}$, then according to \eqref{eq:intpdf} we have the following predicting procedure,
\begin{equation}
\label{eq:eventpred}
	\begin{aligned}
\hat{\tau} &=\int_0^{\infty} s \frac{\partial \phi(s) }{\partial s}\exp(-\phi(s)) ds\\
\centering
\hat{m}&=\underset{\circ}{\mathrm{argmax}} \quad P_m(\circ|\hat{\tau})
	\end{aligned}
\end{equation}
The inconvenience for the time prediction is that the expectation $\hat{\tau}$ often lacks an analytical solution. Consequently, numerical estimation methods become necessary, which not only have the potential to introduce inaccuracies in the prediction results but can also be computationally intensive. To avoid the usage of approximation methods and make fast time prediction, we additionally add a time predictor, realized by a two-layer neural network, i.e.,
\begin{equation}
\label{eq:timepred}
	\begin{aligned}
\bar{\tau}=v_p^T(W_p^1h_i+b_p^1)+b_p^2
	\end{aligned}
\end{equation}
where $W_p^1\in \mathbb{R}^{d\times d}, b_p^1 \in \mathbb{R}^d, v_p^T \in \mathbb{R}^d$ and $b_p^2$ are trainable parameters. Time predictor takes history vector $h_i$ as input and straightforwardly outputs the predicted time for the $(i+1)\mbox{-}th$ event. The whole model  architecture is illustrated in Fig. \ref{fig:model}.

\subsection{Training Loss}
Training a temporal point process model is to maximize its log-likelihood on observed event sequences. Follow the notation in Section \ref{subsec:mtpp}, given the observed sequence $S$, the log-likelihood can be computed as follows,
\begin{equation}
\label{eq:objective}
	\begin{aligned}
\mathcal{L}_l&=\frac{1}{N}\sum_{i=1}^{N}\big[\log P^*(m_i,\tau_i)=\log P_t^*(\tau_i)+\log P_m^*(m_i|\tau_i)\\
&=\log \frac{\partial \phi^*(\tau) }{\partial \tau}\bigg|_{\tau=\tau_i}-\phi^*(\tau_i)+\log P_m^*(m_i|\tau_i)\big]\\
	\end{aligned}
\end{equation}
where superscript $*$ reminds us of the history condition $H_{t_i}$.  The performance on time prediction by the time predictor (TP) in \eqref{eq:timepred} is evaluated by the mean square error, i.e.,
\begin{equation}
\label{eq:timeloss}
	\begin{aligned}
\mathcal{L}_t&=\frac{1}{N}\sum_{i=1}^{N}(\tau_i-\mathrm{TP}(h_{i-1}))^2\\
	\end{aligned}
\end{equation}
We learn the whole model under the multi-task learning framework. Therefore, we construct the final loss as $ \mathcal{L}=-\mathcal{L}_l+\alpha \mathcal{L}_t$. The first term is called negative log-likelihood (NLL) and $\alpha$ is a hyperparameter that controls the loss from time prediction.

\section{Experiment}
\subsection{Datasets}
We evaluate the proposed model over two synthetic datasets Hawkes1, Haweks2, and four real-world datasets Retweets \cite{zuo2020transformer}, SOflow \cite{waghmare2022modeling}, MIMIC \cite{waghmare2022modeling} and Social \cite{Liu2023LinkAwareLP}. Hawkes1, Hawkes2 are simulated by the Hawkes process, where past events will excite future events and the influences decay exponentially over time. Hawkes1 and Hawkes2 have different parameters that control the influences.\footnote{The details can be found at \href{https://github.com/lbq8942/NPP}{https://github.com/lbq8942/NPP}} The four real-world datasets cover different application domains, collecting events from social platform, question answering website, healthcare records and offline social activity, respectively. Each dataset has a number of event sequences and sequences can have different length. These statistics together with the number of event types, mean inter-event time, and its standard deviation (in bracket) are reported in Table \ref{table:dataset}.

\begin{table}
\caption{Dataset statistics.}
\centering
\begin{tabular}{c|cccc}
 \hline
	\multirow{1}{*}{Dataset} & \#Types& \#Seqs& Avg. Length&Interval\\
 \hline
     Hawkes1&9&9494&7.4&0.51(0.55)\\
      Haweks2&9&3960&10.4&0.28(0.50)\\
     Retweet&3&20000 &108.8 &3.98(2.60)\\
         SOflow&22&6633 &72.2&12.72(1.83)\\
    Mimic&75&715 &3.7 &0.41(0.38)\\
        Social&1641&1337 &99.0 &1.18(1.66)\\

\hline
\end{tabular}
\label{table:dataset}
\end{table}
\begin{table*}
\caption{Negative log-likelihood (per sequence ) comparison. The lower the better. The results are averaged by 10 runs and the standard deviation is reported in bracket. The best and second best model on each dataset have been highlighted in bold and underline respectively. OOM means out of memory on 24GB GPU.}
\centering
\scalebox{1.0}{\begin{tabular}{cccccccc}
 \hline
Model &Hawkes1&Hawkes2 &Retweet&SOflow&MIMIC&Social\\
 \hline
  RMTPP&11.3(0.0) &10.1(0.1)&262.8(2.0)&246.0(3.3)&7.3(1.4) &\uline{354.6(26.9)}\\
    NHP&\uline{9.7(0.0)}&9.4(0.0)&253.7(2.3)&238.2(2.9)&7.1(1.3) &OOM\\  
             SAHP &10.2(0.0)&9.6(0.1)&272.8(7.5)&235.4(3.2)&6.9(1.6)&478.7(7.5)\\
         THP &10.4(0.0)&9.8(0.1)& 264.0(1.9)& 237.4(1.4)&6.8(1.7)&488.9(9.8)\\
     UNIPoint&9.9(0.0)&9.4(0.0)&254.9(1.9) &232.3(2.9)&\uline{5.7(1.2)}&OOM \\
   JTPP&10.0(0.0)&\uline{9.3(0.1)} &\textbf{235.4(2.6)}&\uline{225.3(2.7)}&5.8(1.2)&OOM \\
     \sout{FullyNN}&\sout{9.9(0.1)}&\sout{9.7(0.0)}&\sout{238.0(3.2)}&\sout{233.8(6.2)}&\sout{6.7(2.1)}&OOM \\
 \hline
\textbf{\textit{ours}}&\textbf{9.6(0.1)}&\textbf{9.0(0.0)} &\uline{240.4(4.8)}&\textbf{223.9(2.5)}&\textbf{5.6(1.2)} &\textbf{256.5(8.6)}\\
\hline
\end{tabular}}
\label{table:nll}
\end{table*}

\subsection{Baselines}
To evaluate the effectiveness of our model, we compare it against the state-of-the-art models. We categorize the selected baselines into two classes, intensity-based and -free models. We choose RMTPP \cite{du2016recurrent}, NHP \cite{mei2017neural}, SAHP \cite{zhang2020self}, THP \cite{zuo2020transformer} and UNIPoint \cite{soen2021unipoint} for intensity-based models, where SAHP and THP utilize Transformer as the history encoder and the others use the recurrent neural network. There are fewer studies concerning the non-intensity based temporal point process, we choose FullyNN and JTPP \cite{waghmare2022modeling} (extended from LNM \cite{shchur2019intensity} for multivariate TPP) as baselines, they both achieve the best performance in their paper. 

\subsection{Implementation Details}
\label{subsec:imp}
We use Adam algorithm to optimize the parameters, with the learning rate set to 0.001 and the hyperparameter $\alpha$ in the loss function set to 0.01 for all six datasets.  Mini-batch algorithm is adopted for training, with 64 event sequences per batch. Sequences in one batch are padded to the maximum length. To ensure the $\tau$-related parameters in cumulative hazard function network to be positive, we technically take their absolute values each time the parameters are updated.  Each dataset is split into training, validation and testing data, with the proportion being 0.6, 0.2 and 0.2, respectively. The embedding dimension for event types ($d_m$) and the hidden dimension for history vector ($d$) is set to 64 and 128 for dataset Social,  32 and 64 for other datasets. We adopt the early stopping strategy, i.e., stop training when performance on the validation data has not been improved for 10 epochs. The baseline models also apply these setups if exist. The training is conducted on GPU NVIDIA 3090 with 24GB for all models.
\begin{figure}
\centering 
\includegraphics[width=0.5\textwidth]{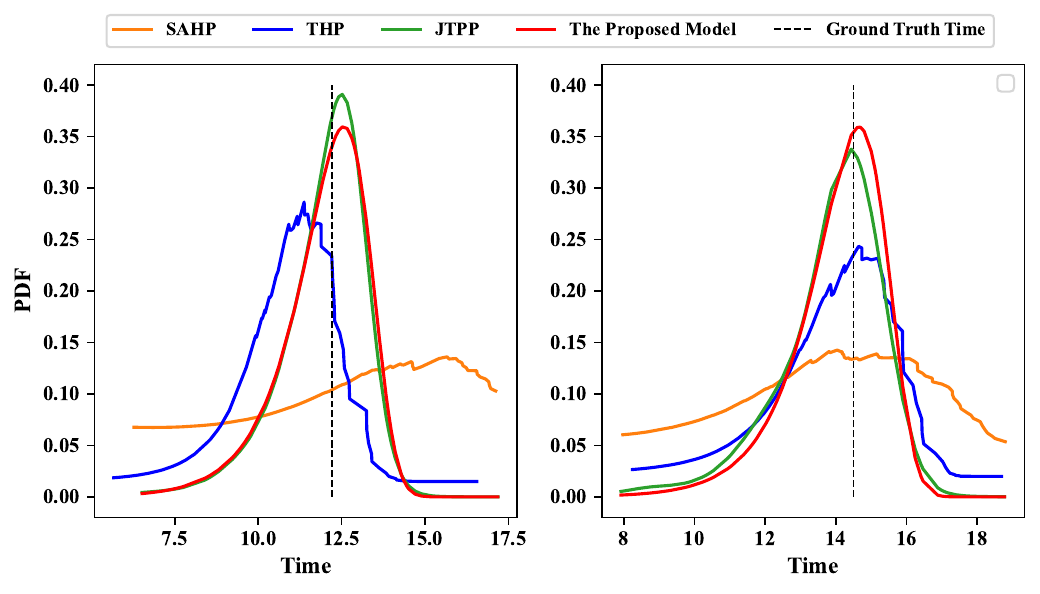}
\caption{Case study: two learned probability density functions on dataset SOflow. The proposed model and JTPP fit the ground truth time more exactly.
}
\label{fig:pdf}
\end{figure}
\begin{table*}
\caption{Event type Prediction Accuracy (weighted F1 score) Comparison. The higher the better. }
\centering
\scalebox{1.0}{\begin{tabular}{ccccccc}
 \hline
Model &Hawkes1&Hawkes2 &Retweet&SOflow&MIMIC&Social\\
 \hline
  RMTPP&37.2(0.6) &22.6(0.6)&59.1(0.0)&31.2(0.2)&63.2(4.5) &\uline{43.1(4.6)}\\
    NHP&38.6(0.3) &\textbf{24.7(0.6)}&59.6(0.1)&31.0(0.8)&63.4(4.8) &OOM\\
         SAHP &36.8(0.3)&22.9(0.2)&58.4(1.2)&31.5(1.0)&65.3(5.8)&21.0(0.2)\\
    THP&38.5(0.5) &23.8(0.4)&59.8(0.2)&31.6(0.8)&65.5(5.6)&29.9(2.0) \\
     UNIPoint&37.8(1.1)&24.0(0.8) &59.7(0.0)&30.7(0.4)&\textbf{66.0(5.2)}&OOM\\
   JTPP&\textbf{39.0}(0.3)&\uline{24.5(0.9)}&\uline{60.0(0.1)}&\textbf{31.8}(0.2)&65.4(5.5)&OOM \\
 \hline
\textbf{\textit{ours}} &\uline{38.7(0.5)}&\textbf{24.7(0.6)}&\textbf{60.2}(0.1)&\uline{31.6(0.2)}&\uline{65.7(4.6)}&\textbf{45.5}(0.5)\\

\hline
\end{tabular}}
\label{table:type}
\end{table*}

\begin{table*}
\caption{Event Time Prediction error (mean absolute error) Comparison. The lower the better.}
\centering
\scalebox{1.0}{\begin{tabular}{ccccccc}
 \hline
Model &Hawkes1&Hawkes2 &Retweet&SOflow&MIMIC&Social\\
 \hline
  RMTPP&0.3755(0.0028) &0.2238(0.0007)&1.9268(0.0067)&4.1043(0.1192)&0.4475(0.0252) &\uline{0.9848(0.0481)}\\
    NHP&0.3704(0.0025) &0.2160(0.0028)&1.8814(0.0056)&5.0963(0.0492)&0.4063(0.0182) &OOM\\
         SAHP &0.3748(0.0045)&0.2154(0.0037)&1.8160(0.0495)&2.9639(0.4214)&0.3335(0.0130)&1.2193(0.0037)\\
    THP&\uline{0.3681(0.0039)}&0.2116(0.0031) & 1.8902(0.0082)&5.1592(0.0315)&0.3546(0.0124)&1.0321(0.0288)\\
     UNIPoint&0.3695(0.0030)&0.2151(0.0021)& 1.8384(0.0030)&4.5776(0.0839)&\textbf{0.2462(0.0046)}&OOM\\
   JTPP&0.3786(0.0268)&\uline{0.1961(0.0252)}& \uline{1.0245(0.0021)}&\uline{1.2557(0.0178)} &0.2755(0.0081)&OOM\\
 \hline
\textbf{\textit{ours}}&\textbf{0.3617(0.0030)}&\textbf{0.1587(0.0028)}&\textbf{1.0158(0.0022)}&\textbf{1.1722(0.0018)} &\uline{0.2511(0.0053)}&\textbf{0.7575(0.0251)} \\

\hline
\end{tabular}}
\label{table:time}
\end{table*}

\subsection{Data Fitting} 
Likelihood indicates how a temporal point process model fits the observed sequences. Table \ref{table:nll} summarizes the negative log-likelihoods (per sequence) on testing data for all the models. Lower negative log-likelihood indicates better performance. FullyNN is denoted with a strikeout, reminding us that this model does not define a valid CHF. As a result, FullyNN has biased results and should not be compared for fairness. Some baselines encounter out of memory on our provided 24GB GPU over dataset Social (a dataset with 1641 event types). We denote this phenomenon as OOM. Table \ref{table:nll} shows that the proposed model achieves the best performance for 5 of 6 datasets, and is the second best on dataset Retweet, which validates our effectiveness. JTPP obtains the second best performance, but can not be applied to datasets with thousands of event types. Both the proposed model and JTPP are non-intensity based, where the log-likelihood can be calculated exactly for training. As a contrast, state-of-the-art intensity-based models such as UNIPoint and NHP require numerical estimation, which hurts their performance though UNIPoint has been theoretically proven to have the universal approximation property. Other baselines including RMTPP, SAHP and THP specify the intensity function to be monotonically decreasing or increasing, thus the flexibility is limited. In our model, the cumulative hazard function is characterized by a neural network, which in principle can be a much more flexible function. A case study is conducted on dataset SOflow to see the learned probability density function. TPP models generate conditional density function at each event position, two of them are shown in Fig. \ref{fig:pdf}. We see that the proposed model and JTPP fit the ground truth time more exactly: the learned density function reaches its peak near the ground truth time and has a smaller entropy.

\subsection{Event prediction}
TPP models encode the past events and generate the conditional distribution for the next event, which can be leveraged for event prediction, including the type prediction and the occurrence time prediction. The predicting procedure for the proposed model has been described in \eqref{eq:eventpred} and \eqref{eq:timepred}. For event type prediction, we use the weighted F1 score to evaluate the performance, and for event time prediction we use the mean absolute error. Higher F1 score and lower time error indicate better performance. The results for these two predictions are reported in Table \ref{table:type} and \ref{table:time}, respectively. Table \ref{table:type} shows that the proposed model outperforms the baselines, though not significantly against JTPP (excluding dataset Social). However, JTPP scales not well to datasets with hundreds and thousands of event types, limiting its application. In Table \ref{table:type}, an interesting phenomenon is that the proposed model seems to have more stable performance, while most baselines perform not well on at least one dataset, e.g., Social. This is because for most baselines, the event type prediction is time-dependent, which means inaccurate time prediction will badly affect the type prediction. The time prediction results in Table \ref{table:time} show that we have the least prediction errors in most datasets while most baselines fall short on the time prediction on dataset SOflow, which as a result causes the error propagation to the type prediction. RMTPP and JTPP both predict the event type by a multi-layer perceptron type predictor (not time-dependent). Hence the time prediction results will not affect their type prediction. We insist the time-dependent type modelling for it's more expressive and could potentially witness more applications: consider the situation where the ground truth time is given or we would like to analyze how type distribution evolves over time. In contrast, the time-independent type predictor can not utilize the timestamps information and thus will be fairly disadvantageous in these cases. For time prediction, we see the proposed model has a significant error reduction compared with baselines on most datasets. Considering both type and time prediction results in Table \ref{table:type} and \ref{table:time}, they highlight the superiority of our proposed model.

\begin{figure*}
\centering 
\includegraphics[width=1.0\textwidth]{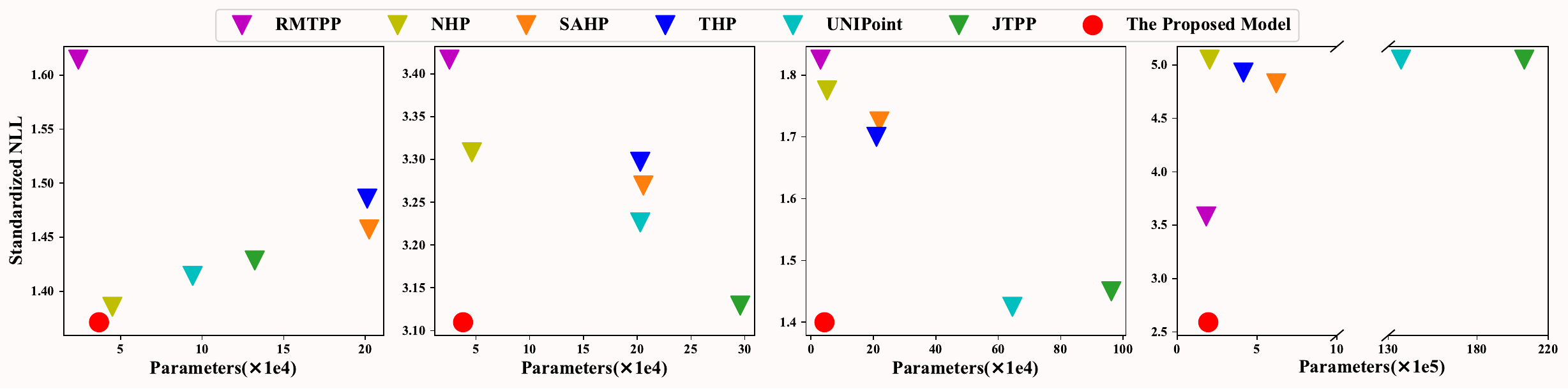}
\caption{The fitting performance and required parameters for baselines and the proposed model on datasets Hawkes1, SOflow, MIMIC and Social. In each plot, models locate in the bottom and left are better.
}
\label{fig:nllpara}
\end{figure*}

\subsection{Model Complexity}
For our mainly focused parameter complexity, the proposed method only models a cumulative hazard function and a time-dependent type predictor for multivariate temporal point process. As a comparison, most baseline methods learn intensity (or density, etc.) functions tailored to each individual event type. These approaches typically involve a substantial number of parameters, particularly in instances where an intricate intensity (or density, etc.) function is designed for flexibility, as exemplified by UNIPoint and JTPP.  Fig. \ref{fig:nllpara} compares the trade-off between the fitting performance and required parameters for all baselines and the proposed model on four datasets. The standardized NLL is the negative log-likelihood in Table \ref{table:nll} divided by the average sequence length (see Table \ref{table:dataset}). NHP, UNIPoint and JTPP can not run on dataset Social due to out of memory, hence we only report their required parameters. Fig. \ref{fig:nllpara} shows that the proposed model not only achieves the best fitting performance but also has almost the least parameters, making it more desirable for multi-type event sequence modelling. In addition to the required parameters, we also report the peak GPU memory usage in Table \ref{table:gpu}. Note that the memory usage is not only affected by the required parameters but also can be affected by some other factors like batch size, sequence length, model architecture, etc. Except for the model architecture, all other factors are controlled to be the same, as introduced in Section \ref{subsec:imp}. Table \ref{table:gpu} again demonstrates the superiority of our proposed model, which enables up to $100 \times$ reduction in memory usage comparing to the state-of-the-art baselines. 
\begin{table}
\setlength{\tabcolsep}{4pt}
\caption{Peak GPU Memory Usage (GB).}
\centering
\scalebox{1.0}{\begin{tabular}{ccccccc}
 \hline
Model &Hawkes1&Hawkes2 &Retweet&SOflow&MIMIC&Social\\
 \hline
   RMTPP&0.01 &0.02&0.05&0.14&0.01 &0.23\\
       NHP &0.26&0.39&0.55&5.80&0.80&OOM\\
          SAHP &0.14&0.25&1.02&6.81&0.17&11.63\\
    THP &0.15&0.25&1.02&6.79&0.18&12.78\\
     UNIPoint &0.55&0.85&0.75&14.61&2.24&OOM\\
   JTPP &0.15&0.23&0.22&3.82&0.59&OOM \\
 \hline
\textbf{\textit{ours}}  &0.03&0.04&0.11&0.32&0.02&0.28\\

\hline
\end{tabular}}
\label{table:gpu}
\end{table}
\begin{table}
\caption{The influence of the hyperparameter $\alpha$ in the loss function, evaluated on dataset Retweet.}
\centering
\scalebox{1.0}{\begin{tabular}{ccccccc}
 \hline
Task&$\alpha$=0.001&$\alpha$=0.01&$\alpha$=0.1&$\alpha$=1\\
 \hline
   data fitting&240.1 &240.4&243.5&271.5\\
       type prediction &60.3&60.2&60.2&60.1\\
          time prediction &1.0146&1.0135&1.0115&1.0123\\
\hline
\end{tabular}}
\label{table:alpha}
\end{table}

\begin{figure}
\centering 
\includegraphics[width=0.45\textwidth]{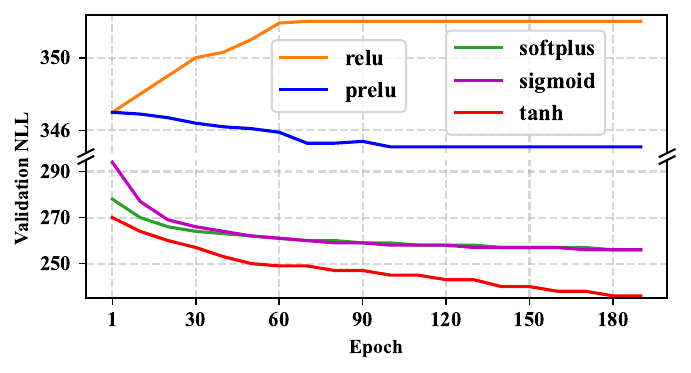}
\caption{Training curves of the proposed model using different activation functions for the cumulative hazard function network, fitted on dataset Retweet.}
\label{fig:activation}
\end{figure}
\subsection{Ablation Studies}
In this part, we investigate the impact of hyperparameter $\alpha$ in the loss function and the impact of the activation function $\sigma_t$ in the cumulative hazard function network. Both of them are conducted on dataset Retweet. The hyperparameter $\alpha$ controls the loss from the time prediction and is set to 0.01 by default. To see how different $\alpha$ affect the model performance, we set it from candidate set $\{0.001, 0.01, 0.1, 1\}$. The results are shown in Table \ref{table:alpha}, from which we can see that the performance on data fitting gets worse when loss from the time prediction more and more dominates the entire loss. But for event prediction, the performance is surprisingly nearly not affected, showing the robustness. 

We show how activation function $\sigma_t$ in the CHF network will dramatically influence the model performance and commonly used activation functions such as $relu$ will lead to inferior performance. Besides $relu$, we also try some other functions that can satisfy the monotonic constraint such as $prelu$, $softplus$ and $sigmoid$, where $prelu$ is defined as  $max(\eta_1 x,x)$ and $softplus$ is defined as $(1/\eta_2) \log(1+\exp(\eta_2 x))$ (both $\eta_1$ and $\eta_2$ are trainable). The investigation results are shown in Fig. \ref{fig:activation}, where we illustrate the learning curves when using different activation functions. The plots show that $tanh$ is the best of those tried activation functions and $relu$, $prelu$ can not fit the data well. We analyze the reason as that $tanh$ can provide more non-linearity and its negative activation can reserve more capacity, both of which enhance the fitting ability.


\section{Conclusion}
In this study, we present a cumulative hazard function based multivariate neural point process model. By modeling CHF, the proposed model enjoys accurate and efficient likelihood evaluation. We design a neural network architecture by which the cumulative hazard function can be well-defined and flexibly modelled. For multivariate modeling, we adopt a parameter-efficient strategy, making our model more scalable. Extensive experiments are conducted to demonstrate the effectiveness of the proposed model: gains better performance while many fewer parameters and memory usage are required.




\bibliographystyle{./IEEEtran}
\bibliography{./IEEEabrv,./ref}


\end{document}